\pdfoutput=1

\documentclass[11pt]{article}

\usepackage[]{ACL2023}

\usepackage{times}
\usepackage{latexsym}

\usepackage[T1]{fontenc}

\usepackage[utf8]{inputenc}

\usepackage{microtype}

\usepackage{inconsolata}

\usepackage{algorithmic} 
\usepackage[ruled,vlined]{algorithm2e} 
\usepackage{multirow} 
\usepackage{graphicx}
\usepackage{amsfonts}
\usepackage{amsmath}
\usepackage{cleveref} 
\usepackage{booktabs}
\usepackage{array}
\usepackage{hyperref}
\usepackage{color, colortbl}

\def\Adapter{\text{adapter}}
\def\Encoder{\text{encoder}}
\def\Bert{\text{Bert}}

\title{ConKI: Contrastive Knowledge Injection for Multimodal Sentiment Analysis}

\author{Yakun Yu $^1$, Mingjun Zhao $^1$, Shi-ang Qi $^1$, Feiran Sun $^2$, Baoxun Wang $^2$, \\\textbf{Weidong Guo} $^2$, \textbf{Xiaoli Wang} $^2$, \textbf{Lei Yang} $^2$, \textbf{Di Niu} $^1$\\
\text{$^1$University of Alberta}\\
\text{$^2$Platform and Content Group, Tencent}\\
\texttt{$^1$\{yakun2, zhao2, shiang, dniu\}@ualberta.ca}
}

\begin{document}
\maketitle
\begin{abstract}
Multimodal Sentiment Analysis leverages multimodal signals to detect the sentiment of a speaker. Previous approaches concentrate on performing multimodal fusion and representation learning based on general knowledge obtained from pretrained models, which neglects the effect of domain-specific knowledge.
In this paper, we propose Contrastive Knowledge Injection (ConKI) for multimodal sentiment analysis, where specific-knowledge representations for each modality can be learned together with general knowledge representations via knowledge injection based on an adapter architecture.
In addition, ConKI uses a hierarchical contrastive learning procedure performed between knowledge types within every single modality, across modalities within each sample, and across samples to facilitate the effective learning of the proposed representations, hence improving multimodal sentiment predictions.
The experiments on three popular multimodal sentiment analysis benchmarks show that ConKI outperforms all prior methods on a variety of performance metrics.
\end{abstract}

\section{Introduction}

Multimodal sentiment analysis (MSA) is the task of mining and comprehending the sentiments of online videos, 
which has many downstream applications, e.g., analyzing the overall opinion from customers about a product, gauging polling intentions from voters \cite{han2021improving, 10.1145/1557019.1557156}, etc. 
Most existing MSA methods focus on developing fusion techniques between modalities. 
The easiest way is to simply concatenate text, video, and audio features as a fused vector for subsequent classification or regression. 
An alternative is to use outer-product, Recurrent Neural Networks (RNNs) or attention-based models to model multimodal interactions \cite{chen2017multimodal, williams-etal-2018-recognizing, zadeh2017tensor,liu2018efficient}. 
More recently, MSA methods for learning effective multimodal representations have emerged constantly \cite{hazarika2020misa, mai2021hybrid, yu2021learning}, ranging from decomposing the representation of each modality to introducing extra constraints in the learning objective.

Although the above methods have led to improvements in MSA performance, they focus on utilizing general knowledge obtained from pretrained models to encode modalities, which is inadequate to identify specific sentiments across modalities.
One possibility to solve this issue is through knowledge injection which can generate specific knowledge to aid the general knowledge for further improving predictions. Many researchers have discovered that injecting knowledge from other sources such as linguistic knowledge, encyclopedia knowledge, and domain-specific knowledge can help enhance existing pretrained language models in terms of knowledge awareness and lead to improved performance on various downstream tasks \cite{wei2021knowledge, lauscher2020common, wang2021k}.

In this paper, we propose ConKI, a Contrastive Knowledge Injection framework, to learn both pan-knowledge representations and knowledge-specific representations to boost MSA performance. We argue that a unimodal representation can consist of a pan-knowledge representation (given by a pretrained model like BERT \cite{Devlin2019BERTPO}) 
and a knowledge-specific representation (injected from relevant external sources).
Specifically, ConKI uses a pretrained BERT model to extract textual pan-knowledge representations and uses two randomly initialized transformer encoders to generate acoustic and visual pan-knowledge representations, respectively. In the meantime, it applies a knowledge injection model named adapter, onto each modality to yield knowledge-specific representations. Both pan- and specific-knowledge representations are fused first within each modality and then across modalities, before the fused features are used for sentiment prediction. 
We further propose a hierarchical contrastive learning procedure performed between knowledge types within every single modality, across modalities within each sample, and across samples, to facilitate the learning of these representations in ConKI.

The main contributions of this work can be summarized as follows:
\begin{itemize}
    \item We propose ConKI, a Contrastive Knowledge Injection framework for multimodal sentiment analysis. ConKI aims to boost model performance through external knowledge injection from other datasets and hierarchical contrastive learning, which is proved better than simply fine-tuning with external datasets.
    \item We propose hierarchical contrastive learning that uses a unified contrastive loss to disentangle the pan-knowledge representations from the specific-knowledge representations since they belong to different knowledge domains and should complement each other.  
    \item We conduct extensive experiments on three popular benchmark MSA datasets and attain results that are superior to the existing state-of-the-art MSA baselines on all metrics, demonstrating the effectiveness of the proposed methods in ConKI. 
\end{itemize}

\section{Related Work}
In this section, we discuss related research in multimodal sentiment analysis, knowledge injection, and contrastive learning. 

\subsection{Multimodal Sentiment Analysis}
Research on MSA mainly focuses on multimodal fusion and representation learning.
For multimodal fusion, existing methods are typically divided into early fusion and late fusion techniques.
Early fusion refers to joining multimodal inputs into a single feature before single-model encoding.
For example, 
\citet{williams-etal-2018-recognizing} concatenate initial input features and then use LSTM to capture the temporal dependencies in the sequence. 
 On the contrary, late fusion learns unimodal representations via separate models and fuses them in a later stage for inference. 
 \citet{zadeh2017tensor} introduce a tensor fusion network that first encodes each modality with corresponding sub-networks and then models the unimodal, bimodal, and trimodal interactions by a three-fold Cartesian product. 
For representation learning methods, 
\citet{hazarika2020misa} propose to project each modality into a modality-invariant and modality-specific representation.   
Different from the above work, we propose to decompose each modality into two representations based on knowledge types. 
Both representations can complement each other, leading to a richer unimodal representation.

\subsection{Knowledge Injection}
Injecting knowledge into pretrained language models (PLMs) has been proven to outperform vanilla pretrained models on various NLP tasks \cite{wei2021knowledge, wang2021k, tian2020skep, ke2020sentilare, lin2019kagnet, wang2021kepler}. 
Adapters are commonly used as a knowledge injection model plugged outside or inside of PLMs. For instance,
\citet{wang2021k} infuse factual knowledge from Wikidata \cite{vrandevcic2014wikidata} and linguistic knowledge from web text to RoBERTa \cite{liu2019roberta} via two kinds of adapters.
In this work, we build different adapters for different modalities, not limited to text, to learn specific multimodal knowledge from an external dataset for the downstream task.  
To the best of our knowledge, we are the first to explore knowledge injection in the multimodal domain.

\begin{figure*}[!htp]
  \centering
  \includegraphics[width=\textwidth]{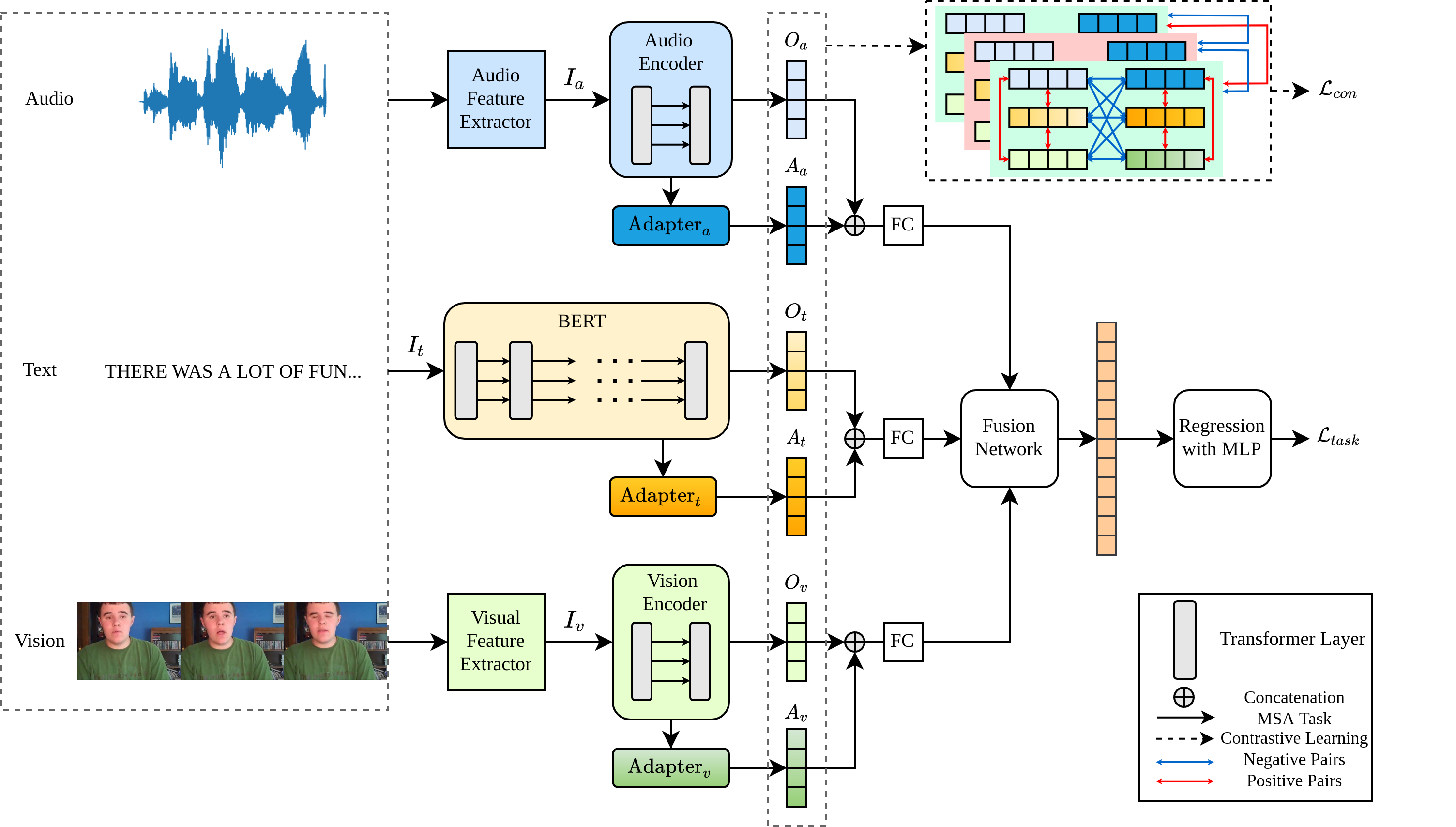}
  \caption{The overall architecture of ConKI. The solid and dashed arrows represent the procedure of the main MSA task and the hierarchical contrastive learning subtask, respectively. Inside the contrastive learning procedure, cyan and pink boxes illustrate samples that fall in different sentiment score intervals.}
  \label{fig:framework}
\end{figure*}

\subsection{Contrastive Learning}
Contrastive learning (CL) aims to learn effective representations such that positive pairs of samples are close while negative pairs of samples are far apart \cite{Liu2021LearningAF, li2020prototypical, chen2020simple, khosla2020supervised, he2020momentum}. Existing works can be divided into two categories: self-supervised CL \cite{akbari2021vatt, chen2020simple, chen2020big, he2020momentum, you2020graph, tao2020self} and supervised CL \cite{khosla2020supervised, mai2021hybrid}. The difference between them is whether the label information is used to form positive/negative pairs. 
For example, 
\citet{khosla2020supervised} propose supervised CL to pull samples of the same class together and push samples from different classes away. 
In our work, we design contrastive pairs in finer granularity. That is, we consider contrasts between knowledge types, between modalities, and across samples.

\section{Method}
In this section, we explain the Contrastive Knowledge Injection framework (ConKI) in detail. 
The goal of ConKI is to generate pan- and specific-knowledge modality representations via knowledge injection and hierarchical contrastive learning.
Knowledge injection intends to obtain knowledge-specific representations that could complement
the pan-knowledge representations offered by pretrained models. 
Hierarchical contrastive learning further optimizes these knowledge-specific and pan-knowledge representations by considering contrasts between knowledge types, modalities, and samples.

\subsection{Problem Definition}
The task of multimodal sentiment analysis (MSA) is to detect sentiments in videos based on multimodal signals, including text ($t$), vision ($v$), and audio ($a$) modalities.
These signals are represented as sequences of low-level features, i.e., $I_t \in \mathbb{R}^{l_t\times d_t}$, $I_v \in \mathbb{R}^{l_v\times d_v}$,
and $I_a \in \mathbb{R}^{l_a\times d_a}$, respectively. Here $l_{m \in \{t, v, a\}}$ denotes the length of the sequence for each modality,
while $d_{m \in \{t, v, a\}}$ denotes the corresponding feature vector dimension. The detail for acquiring these features is described in Appendix \ref{Imple_details}.
Given these sequences $I_{m \in \{t, v, a\}}$, the primary task is to make accurate predictions on the sentiment intensity by extracting and fusing higher-level multimodal information. 

\subsection{Overall Architecture}

Figure \ref{fig:framework} shows the overall architecture of ConKI. We first process raw multimodal input to low-level features $I_{m \in \{t, v, a\}}$ with their corresponding feature extractors and tokenizers. 
Then we encode $I_m$ into knowledge-specific representations (i.e., $A_m$) generated by some adapters and pan-knowledge representations (i.e., $O_m$) generated by pretrained encoders. The text encoder is from publicly-available pretrained backbones like BERT \cite{Devlin2019BERTPO}, and the vision/audio encoder is a designed model with random initialization since there is no suitable backbone that is pretrained by the above low-level features. 
After generating the knowledge-specific and pan-knowledge representations, ConKI is trained simultaneously with two different tasks on the downstream target dataset -- the primary MSA regression task and the contrastive learning subtask. 

\begin{figure}[t]
    \centering
  \includegraphics[width=\columnwidth]{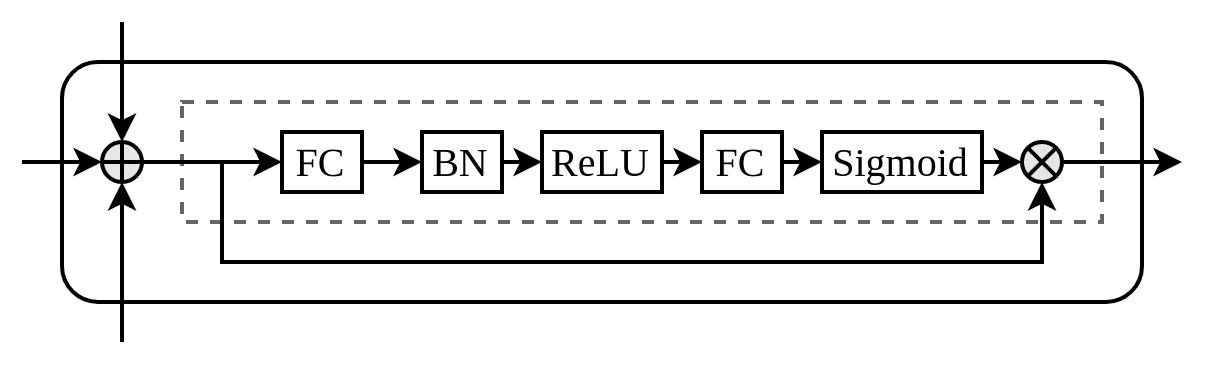}
  \caption{The Fusion Network. The fusion module marked in the dashed box is used to get the weighted fused embedding. $\bigotimes$ means element-wise multiplication.} 
  \label{fig:fusion}
\end{figure}

For the MSA task, we concatenate the knowledge-specific representation and pan-knowledge representation of each modality before feeding them into a fully-connected (FC) layer for inner-modality fusion. We then design a fusion network that consists of a concatenation layer and a fusion module for multi-modality fusion, as shown in Figure \ref{fig:fusion}. The fused representations are passed into a multilayer perceptron (MLP) network to produce the sentiment predictions, $\hat{y}$.

For the subtask of hierarchical contrastive learning, we carefully construct the negative and positive sample pairs at the knowledge level, modality level, and sample level.
The intuition of our pairing policy is as follows.
We expect $A_m$ and $O_m$ to capture different knowledge, so we disentangle them and make them complement each other to get richer modality representations by knowledge-level contrasts. 
Since a video's sentiment is determined by all modalities, we learn the commonalities among the six representations by modality-level contrasts.
Besides, videos that express close sentiments should share some correlations. We capture the correlations by sample-level contrasts to help further learn the commonalities among samples under close sentiments. 
By integrating these hierarchical contrasts, ConKI is able to catch full dynamics among representations which can significantly benefit the main MSA task.

\subsection{Encoding with Knowledge Injection}

We encode each modality into a pan-knowledge representation via the pretrained encoders and a knowledge-specific representation via the adapters. 

\textbf{Pan-knowledge representations.}
We use the pretrained BERT \cite{Devlin2019BERTPO} to encode the input sentence for the text modality. The pooled output vector in the last layer is extracted as the whole sentence representation $O_t$: 
\begin{equation} \label{eq1}
    O_t, H_t = \Bert(I_t; \theta_t^\Bert) \ ,
\end{equation}
where $H_t$ denotes the hidden states of all layers.
For audio and vision modalities, we employ encoders of stacked transformer layers \cite{vaswani2017attention} to  capture the temporal features $O_m$:
\begin{equation} \label{eq2}
    O_m, H_m = \text{Encoder}(I_m; \theta_m^\Encoder), \; m\in{\{v, a\}} \ .
\end{equation}
Here, $O_t$, $O_a$, and $O_v$ are regarded as three pan-knowledge representations since they mainly contain general knowledge such as the generic facts encoded by BERT \cite{Devlin2019BERTPO} pretrained on big text data.  

\begin{figure}[t]
    \centering
  \includegraphics[width=0.90\columnwidth]{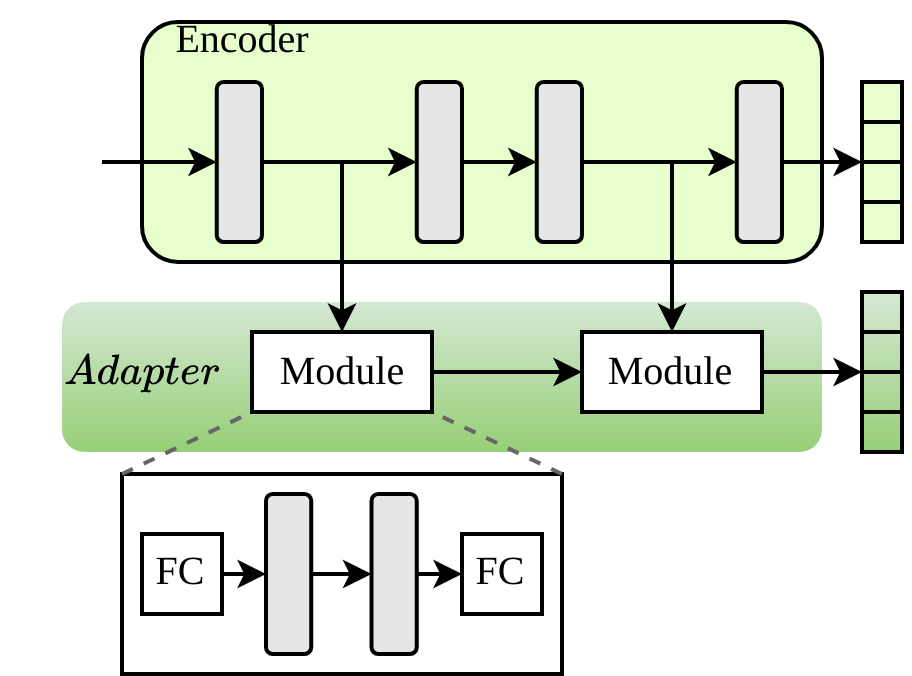}
  \caption{Adapter and its connection with the backbone encoder.}
  \label{fig:adapter}
\end{figure}

\textbf{Knowledge-specific representations.}
We infuse specific domain knowledge from external multimodal sources through knowledge injection models (adapters). The adapter is commonly used in natural language processing (NLP) to enhance existing pretrained language models' knowledge awareness \cite{wei2021knowledge}. The outputs of adapters are taken as knowledge-specific representations.
Specifically, the adapter for each modality is plugged outside of the respective pretrained encoder, as shown in Figure \ref{fig:adapter}. It consists of multiple modules with the same sandwich structure: two FC layers with two transformer layers in between. Each module can be inserted before any transformer layers of the backbone models (encoders), e.g., the second and fourth transformer layers in Figure \ref{fig:adapter}. Therefore, each module takes the intermediate layers' hidden states of the pretrained encoder and the output of the previous adapter module as input. The output of the adapter is denoted as $A_m$, where
\begin{equation} \label{eq3}
    A_m = \text{Adapter}(H_m; \theta_m^\Adapter), \; m\in{\{t, v, a\}} \ .
\end{equation}

With the objective of learning specific multimodal sentiment knowledge, we pretrain one adapter for each modality, i.e., $\text{Adapter}_t$, $\text{Adapter}_a$ and $\text{Adapter}_v$, concurrently using an external dataset while keeping the pretrained encoders frozen. Since the external dataset we select is also from the multimodal sentiment domain, the pretraining task remains the MSA task. That is, we pretrain adapter parts in Figure \ref{fig:framework} with only the MSA task on the external dataset, then utilize the pretrained adapters to produce knowledge-specific representations $A_m$ for the downstream target task that includes both the MSA task and the hierarchical contrastive learning subtask. 
Algorithm \ref{alg:conki} summarizes this pretraining procedure of adapters.

\begin{algorithm}[!t]
\caption{Learning Procedure of ConKI}\label{alg:conki}
\SetAlgoLined
\textbf{Stage 1: Adapter Pretraining}\\
\textbf{Input:} External dataset $\mathcal{E}$, its corresponding features $I_m$ and labels $y$,\\
\textbf{Output:} Pretrained adapters $\{\theta_m^\Adapter \mid m\in\{t, v, a\}\}$ \\
\For{each training epoch} 
{\For{batch $\{(I_t^{i}, I_v^{i}, I_a^{i})\}_{i=1}^{|B|}$ from $\mathcal{E}$}{
\nl Encode $I_m^i$ to $O_m^i$ and $A_m^i$ via Eq. (1-3)\\
\nl Inner-modality fusion: 
    $F_m^i = FC([O_m^i; A_m^i])$,
where $[\cdot;\cdot]$ denotes the concatenation of two vectors\\ 
\nl Multi-modality fusion:
    $F^i = FN(F_t^i, F_v^i, F_a^i)$,
where $FN$ is the fusion network\\
\nl Compute the predictions using $\hat{y^i} = MLP(F^i)$ \\ 
\nl Compute $\mathcal{L}_{task}$ via Eq. (\ref{eq9}) \\
\nl Update parameters except $\{\theta_t^\Bert, \theta_m^\Encoder \mid m\in\{v, a\}\}$
}
Save $\{\theta_m^\Adapter \mid m\in\{t, v, a\}\}$ when reaching the best validation result
}
\textbf{Stage 2: Downstream Fine-tuning}\\
\textbf{Input:} Target dataset $\mathcal{D}$,
                its corresponding features $I_m$, its labels $y$,
                and the pretrained adapters\\
\textbf{Output:} Predictions $\hat{y}$ \\
\For{each training epoch} 
{\For{batch $\{(I_t^{i}, I_v^{i}, I_a^{i})\}_{i=1}^{|B|}$ from $\mathcal{D}$}{
Perform Steps $1-4$ in Stage 1\\
Compute $\mathcal{L}$ via Eq. (\ref{eq10})\\
Update parameters except $\{\theta_m^\Adapter \mid m\in\{t, v, a\}\}$
}}
\end{algorithm}

\subsection{Hierarchical Contrastive Learning} \label{cl}
In our framework, we propose a hierarchical contrastive learning method to enhance the learned representations by considering the following four aspects in a batch $B$:
\begin{itemize}
\itemsep0em
    \item For a single video sample $i$, all the modalities share common motives of the speaker that determine the overall sentiment. 
    The pan-knowledge representations of different modalities are expected to represent similar meanings and thus need to be pulled closer to each other. And the same applies for knowledge-specific representations. This intuition leads to the construction of \textit{intra-sample} positive pairs:
    \begin{equation*} \label{eq4}
    \begin{aligned}
        \mathcal{P}_{1}^i &= \{(O_m^i, O_n^i), (A_m^i, A_n^i) \mid \\ 
        &m, n \in {\{t, v, a\}} \And m\neq n \And i \in B\} \ ;
    \end{aligned}
    \end{equation*}
    \item The pan-knowledge representations and the knowledge-specific representations should be disentangled from each other since they belong to different knowledge domains and are designed to complement each other. This exists inside each sample ($i$ and $j$ represent the same sample) as well as across samples in the batch ($i$ and $j$ represent two different samples). Therefore, we can build the \textit{inter-knowledge} negative pairs within a batch:
    \begin{equation*} \label{eq5}
    \begin{aligned}
         \mathcal{N}_{1}^i = \{&(O_m^i, A_n^j) \mid \\
         & m, n \in {\{t, v, a\}} \And i, j \in B\} \ ;
    \end{aligned}
    \end{equation*}
    \item For two arbitrary samples $i$ and $j$ having close sentiments, i.e., their sentiment scores can be rounded to the same integer,  six representations of sample $i$ (i.e., $O_m^i$ and $A_m^i$) should be close to  the corresponding representations of sample $j$ (i.e., $O_n^j$ and $A_n^j$). Note that the subscripts $m$ and $n$ represent the modality for sample $i$ and $j$, respectively. We then form the \textit{inter-sample} positive pairs as
    \begin{equation*} \label{eq6}
    \begin{split}
    \mathcal{P}_{2}^i = \{(O_m^i, O_n^j), (A_m^i, A_n^j) |m, n \in {\{t, v, a\}}\\
    \And r(y^i)=r(y^j) \And i, j\in B \And i\neq j\} \ ,
    \end{split}
    \end{equation*}
    where $y^i$ denotes the ground-truth of sample $i$, and $r(\cdot)$ stands for the round function;
    \item Except for the pairs derived from the above three aspects, the remaining pairs with sample $i$ in the same batch are set as negative pairs $\mathcal{N}_{2}^i$. Please refer to Appendix \ref{pair} for a more detailed pairing policy. 
\end{itemize}

Specifically, our hierarchical contrastive loss $\mathcal{L}_{con}$ is computed by
\begin{equation*} 
\label{eq8}
\begin{aligned}
    &\mathcal{L}_{con} =  \sum_{i \in B}\frac{-1}{|\mathcal{P}^i_1\cup\mathcal{P}^i_2|} \ \times \\
    & \qquad \sum_{(p,q)\in\mathcal{P}^i_1\cup\mathcal{P}^i_2} \log \frac{f((p, q))}{\sum_{(p',q')\in \text{All Pairs}}f((p', q'))} \ ,
\end{aligned}
\end{equation*}
where,
\begin{equation*}
    f((p, q)) = \exp(\frac{p}{\|p\|_2}\cdot \frac{q}{\|q\|_2} \cdot \frac{1}{\tau})\ ,
\end{equation*}
\begin{equation*}
    \text{All Pairs} = \mathcal{P}^i_1\cup\mathcal{P}^i_2\cup\mathcal{N}^i_1\cup\mathcal{N}^i_2 \ .
\end{equation*}
In the above equation, $|\mathcal{P}^i_1\cup\mathcal{P}^i_2|$ means the number of positive pairs with sample $i$ in a batch $B$, $(\cdot, \cdot)$ denotes a pair in the corresponding set, e.g., $(O_t^i, O_v^i)$, and 
$\tau$ is a scalar temperature parameter. 

The rationale behind this hierarchical contrastive learning subtask is as follows. First, we capture the commonalities across the three modalities within each knowledge type of each sample to reduce the modality gaps under a shared motive. Second, we model the commonalities across samples of close sentiments within each knowledge type to reduce the sample gaps. Third, we capture the differences between the pan-knowledge representations and the knowledge-specific representations in each sample which results in a complementary effect of the two knowledge types of representations. Last but not least, we capture the differences across samples of different sentiments within each knowledge type in order to learn the dynamics of different sentiment intervals.

\subsection{Training Procedure}
Given the ground truth $y$ and the predictions $\hat{y}$, we can calculate the main MSA task loss by the mean squared error:
\begin{equation} \label{eq9}
    \mathcal{L}_{task} = \frac{1}{|B|}\sum_{i}^{|B|}(\hat{y}^i-y^i)^2 \ ,
\end{equation}
where $|B|$ is the number of samples in a batch.
 
ConKI adopts the learning regime of pretraining followed by fine-tuning. We first pretrain the adapters in ConKI with $\mathcal{L}_{task}$ using an external dataset while fixing the model parameters of the pretrained backbones, considering ConKI only encodes specific knowledge in adapters which have much fewer trainable parameters compared to backbones. Then we fine-tune ConKI with the downstream target dataset by optimizing the overall loss $\mathcal{L}$:
\begin{equation} \label{eq10}
    \mathcal{L} = \mathcal{L}_{task} + \lambda \mathcal{L}_{con} \ ,
\end{equation}
where $\lambda$ is a hyperparameter that balances the MSA task loss and the hierarchical contrastive loss. Algorithm \ref{alg:conki}
shows the full training procedure of ConKI.

\begin{table}[t]
  \resizebox{\columnwidth}{!}{\begin{tabular}{cccccl}
    \toprule
    Dataset & \#Train & \#Valid  & \#Test & \#Total \\
    \midrule
    CMU-MOSI   & 1284       & 229       & 686       & 2199      \\
    CMU-MOSEI   & 16326       & 1871       & 4659       & 22856      \\
    SIMS &1368 &456 &457 &2281 \\
  \bottomrule
\end{tabular}}
\caption{The statistics of CMU-MOSI, CMU-MOSEI and SIMS.}
\label{tab:datasets}
\end{table}


\begin{table*}[t]
\resizebox{\textwidth}{!}{\begin{tabular}{lcccccccccc}
\toprule
\multirow{2}{*}{Models$^\ast$} & \multicolumn{5}{c}{CMU-MOSI}                                                                  & \multicolumn{5}{c}{CMU-MOSEI}                                                                  \\
\cmidrule(lr){2-6}
\cmidrule(lr){7-11}
                        & MAE            & Corr           & Acc-7          & Acc-2                & F1                   & MAE            & Corr           & Acc-7          & Acc-2                & F1                   \\ \midrule
TFN$^\dagger$                     & 0.901          & 0.698          & 34.9           & -/80.8               & -/80.7               & 0.593          & 0.700          & 50.2           & -/82.5               & -/82.1               \\
LMF$^\dagger$                     & 0.917          & 0.695          & 33.2           & -/82.5               & -/82.4               & 0.623          & 0.677          & 48.0           & -/82.0               & -/82.1               \\
MulT$^\dagger$                    & 0.861          & 0.711          & -              & 81.5/84.1            & 80.6/83.9            & 0.580          & 0.703          & -              & -/82.5               & -/82.3               \\
ICCN$^\dagger$                    & 0.862          & 0.714          & 39.0           & -/83.0               & -/83.0               & 0.565          & 0.713          & 51.6           & -/84.2               & -/84.2               \\
MISA$^\dagger$                    & 0.804          & 0.764          & -              & 80.79/82.10          & 80.77/82.03          & 0.568          & 0.724          & -              & 82.59/84.23          & 82.67/83.97          \\
MAG-BERT$^\dagger$                & 0.727          & 0.781          & 43.62          & 82.37/84.43          & 82.50/84.61          & 0.543          & 0.755          & 52.67          & 82.51/84.82          & 82.77/84.71          \\
Self-MM$^\dagger$                 & 0.712          & 0.795          & 45.79          & 82.54/84.77          & 82.68/84.91          & 0.529          & 0.767          & 53.46          & 82.68/84.96          & 82.95/84.93          \\
HyCon$^\ddagger$                   & 0.713          & 0.790          & 46.6           & -/85.2               & -/85.1               & 0.601          & 0.776          & 52.8           & -/85.4               & -/85.6               \\
MMIM$^\dagger$                    & 0.700          & 0.800          & 46.65          & 84.14/86.06          & 84.00/85.98          & \textbf{0.526} & 0.772          & 54.24 & 82.24/85.97          & 82.66/85.94          \\ \midrule
ConKI                   & \textbf{0.681} & \textbf{0.816} & \textbf{48.43} & \textbf{84.37/86.13} & \textbf{84.33/86.13} & 0.529          & \textbf{0.782} & \textbf{54.25}          & \textbf{82.73/86.25} & \textbf{83.08/86.15} \\ \bottomrule
\end{tabular}}
\caption{Results on CMU-MOSI and CMU-MOSEI. In Acc-2 and F1, the left of the ``/'' corresponds to ``negative/non-negative'' and the right corresponds to ``negative/positive''. $^\ast$: all models use BERT as the text encoder; $^\dagger$:from \cite{han2021improving}; $^\ddagger$:from \cite{mai2021hybrid}. Best results are marked in bold.}
\label{tab-MOSI}
\end{table*}

\section{Experiments}
In this section, we present some experimental details, including datasets, evaluation metrics, baseline models, and experimental results. 
The implementation details are shown in Appendix \ref{Imple_details}.
\subsection{Datasets and Metrics}
We conduct experiments on three publicly available benchmark datasets in MSA: CMU-MOSI \cite{zadeh2016mosi}, CMU-MOSEI \cite{Zadeh2018MultimodalLA} and SIMS \cite{yu-etal-2020-ch}. Table \ref{tab:datasets} shows the statistics of the datasets. Appendix \ref{datasets} describes the details of these datasets.

Following the previous works \cite{sun2020learning, rahman2020integrating, hazarika2020misa, yu2021learning, mai2021hybrid, han2021improving, yu-etal-2020-ch}, we report our experimental results in two forms: regression and classification. For regression, we report mean absolute error (MAE) 
and Pearson correlation (Corr). 
For classification, we report binary classification accuracy (Acc-2) and F1 score. Specifically, for CMU-MOSI and CMU-MOSEI datasets, we calculate Acc-2 and F1 scores in negative/positive (zero excluded) and non-negative/positive (zero included) settings as well as seven-class classification accuracy (Acc-7) which shows the percentage of predictions that correctly classified into the same interval of seven intervals between $-3$ and $+3$.
Higher values indicate better performance for all metrics except for MAE.

\subsection{Baselines}
We compare ConKI with the following state-of-the-art baseline models in MSA:
TFN \cite{zadeh2017tensor}, LMF \cite{liu2018efficient}, MulT \cite{tsai2019multimodal}, ICCN \cite{sun2020learning}, MISA \cite{hazarika2020misa}, MAG-BERT \cite{rahman2020integrating}, Self--MM \cite{yu2021learning}, HyCon \cite{mai2021hybrid}, and MMIM \cite{han2021improving}. The details of these baseline models are shown in Appendix \ref{baselines}.

\subsection{Results}
In accordance with previous work, we run our model five times under the same hyper-parameter settings and report the average performance of all metrics in Table \ref{tab-MOSI} and Table \ref{tab-SIMS}. We can observe from these tables that ConKI yields better or competitive results to a range of baseline models on CMU-MOSI, CMU-MOSEI, and SIMS. Specifically, ConKI outperforms all state-of-the-art baseline models in all metrics on CMU-MOSI and SIMS as well as in Corr, Acc-7, Acc-2, F1 scores on CMU-MOSEI. It also achieves closed performance to the best baseline model in MAE on CMU-MOSEI.
\begin{table}[ht]
\centering
\resizebox{0.9\columnwidth}{!}{\begin{tabular}{ccccc}
     \toprule
    Models    & MAE            & Corr           & Acc-2                   & F1             \\
    \midrule
    TFN      & 0.488          & 0.496          & 75.27          & 75.56          \\
    LMF      & 0.487          & 0.502          & 75.36          & 75.78          \\
    MulT     & 0.485          & 0.504          & 75.62          & 75.84          \\
    MISA     & 0.472          & \textbf{0.542} & 75.49          & 75.85          \\
    MAG-BERT & 0.553          & 0.242          & 71.43          & 63.68          \\
    Self-MM  & 0.458          & 0.535          & 77.37          & 77.54          \\
    MMIM     & 0.607          & --             & 69.37          & 58.00          \\ \midrule
    ConKI    & \textbf{0.454} & \textbf{0.542} & \textbf{77.94} &\textbf{78.17} \\
    \bottomrule
\end{tabular}}
\caption{Results on SIMS. All baseline model codes are from \url{https://github.com/thuiar/MMSA}.}
\label{tab-SIMS}
\end{table}

It is notable that the MAE of ConKI on CMU-MOSI outperforms the best baseline model MMIM by around 0.02, which shows ConKI is able to learn effective representations for the MSA task since MAE is the most commonly used evaluation metric in regression tasks. 
ConKI also presents an excellent performance in the Corr scores on both CMU-MOSI and CMU-MOSEI datasets. The possible reasoning behind this excellent performance is that ConKI uses contrastive learning for recognizing the samples under different sentiments, which could lead to effective ranking results among samples and thus produce a higher Corr score \cite{swinscow2002statistics}.

Furthermore, Acc-7 of ConKI on CMU-MOSI surpasses the best baseline by 1.78. Though performing classification, especially seven-class classification, is difficult in a regression task, ConKI successfully leverages the contrasts across samples that are classified into seven intervals (by the round function described in Section \ref{cl}) to model the sample dynamics, which brings a great improvement to Acc-7 and Acc-2, demonstrating the efficacy of ConKI in representation learning for MSA. In addition, ConKI shows excellent F1 scores on all datasets, which endorse its potential in real-world applications since F1 is valuable for evaluating imbalanced datasets.

\subsection{Ablation Study}
\begin{table}[t]
    \resizebox{\columnwidth}{!}{\begin{tabular}{cccccc}
     \toprule
     Models &MAE &Corr &Acc-7 &Acc-2 &F1\\
    \midrule
    V+A     &1.408 &0.248 &18.72  &55.71/54.33  &54.37/53.24\\
    T+A     &0.700 &0.799 &48.22 &82.45/84.18 &82.38/84.16\\
    T+V       &0.718 &0.798 &45.45 &82.97/84.88  &82.89/84.86\\
    \midrule
    ConKI &\textbf{0.681} &\textbf{0.816} &\textbf{48.43} &\textbf{84.37/86.13} &\textbf{84.33/86.13}\\
    \bottomrule
\end{tabular}}
\caption{Ablation results when using different modalities.}
    \label{tab-ab2}
\end{table}

We first conduct an ablation study about modalities, as shown in Table \ref{tab-ab2}. We can observe that the inclusion of all three modalities significantly improves the performance of ConKI. 

To show the benefits of the proposed knowledge injection and hierarchical contrastive learning in ConKI, we conduct a series of ablation experiments on CMU-MOSI, as shown in Table \ref{ablation} and Table \ref{tab-ab1}. 
ConKI mainly includes four components: the use of the external dataset (C1), adapters for knowledge injection (C2), pretrained encoders for pan-knowledge (C3), and hierarchical contrastive learning (C4).
Table \ref{ablation} shows that C1 provides advantages by comparing w/o C1 and ConKI. Similarly, C2 provides benefits by comparing w/o C2 and ConKI. C3 is beneficial by comparing w/o C3 and ConKI. C4 is beneficial by comparing w/o C4 and ConKI.
 
Since the spotlight in our hierarchical contrastive learning is the contrasts between knowledge types, we also compare our model with the model w/o $\mathcal{N}_1$ trained with $\mathcal{L}_{con}$ but without negative pairs $\mathcal{N}_1$, i.e., without disentangling the pan knowledge and specific knowledge. We can conclude that learning differentiated pan-knowledge and knowledge-specific representations is essential in our hierarchical contrastive learning. 
To better understand the learned pan- and specific-knowledge representations by our hierarchical contrastive learning, we visualize and analyze these representations in Appendix \ref{visual}.
\begin{table}[t]
\resizebox{\columnwidth}{!}{\begin{tabular}{lccccc}
\toprule
Models   & MAE            & Corr           & Acc-7          & Acc-2                & F1                   \\ \midrule
ConKI    & \textbf{0.681} & \textbf{0.816} & \textbf{48.43} & \textbf{84.37/86.13} & \textbf{84.33/86.13} \\
w/o C1 & 0.734          & 0.794          & 43.56          & 82.39/84.21          & 82.35/84.22          \\ 
w/o C2 &0.753 &0.789 &43.50 &82.16/83.84 &82.15/83.89 \\
w/o C3 &0.710 &0.811 &44.75 &83.29/84.82 &83.23/84.80 \\\midrule


w/o C4 & 0.683          & 0.815          & 48.05          & 84.23/86.01          & 84.16/85.99          \\
w/o $\mathcal{N}_1$   & 0.689          & 0.812          & 47.90          & 83.88/85.76          & 83.8/85.74           \\ \bottomrule
\end{tabular}}
\caption{Ablation results of ConKI's components on CMU-MOSI.}
\label{ablation}
\end{table}

To further examine if our performance gain is from the external dataset instead of the proposed knowledge injection and contrastive learning technique, we compare our model with the state-of-the-art baseline models which are fine-tuned by the external dataset. The results from Table \ref{tab-ab1} show that ConKI still outperforms those baseline models even though they are trained with the external dataset.

Therefore, our gain from ConKI is not solely from adding more data, but from knowledge injection with multi-step transfer learning.   
Considering the size of CMU-MOSEI is much larger than CMU-MOSI, injecting CMU-MOSEI's knowledge into CMU-MOSI thus has more effects on the downstream task than injecting CMU-MOSI into CMU-MOSEI, as shown in Table \ref{tab-MOSI}. 

\begin{table}[t]
    \resizebox{\columnwidth}{!}{\begin{tabular}{cccccc}
     \toprule
     Models &MAE &Corr &Acc-7 &Acc-2 &F1\\
    \midrule
    MISA   &0.711 &0.804 &44.96 &82.04/83.99 &81.98/84.0\\
    Self-MM     &0.712 &0.798 &45.51 &83.59/85.18 &83.47/85.12\\
    MMIM &0.716 &0.791 &45.42 &81.81/83.54 &81.67/83.46\\
    \midrule
    ConKI &\textbf{0.681}$^\dagger$ &\textbf{0.816}$^\dagger$ &\textbf{48.43}$^\dagger$ &\textbf{84.37$^\dagger$/86.13$^\dagger$} &\textbf{84.33$^\dagger$/86.13$^\dagger$}\\
    \bottomrule
\end{tabular}}
\caption{Ablation results when introducing CMU-MOSEI as an external dataset on CMU-MOSI. $^\dagger$ means the corresponding result is significantly better than SOTA with $p$-value $< 0.05$ based on paired $t$-test.}
    \label{tab-ab1}
\end{table}

\section{Conclusion}
In this paper, we present ConKI, a Contrastive Knowledge Injection framework for multimodal sentiment analysis, which learns knowledge-specific representations along with pan-knowledge representations via knowledge injection and hierarchical contrastive learning.
ConKI utilizes the pretrained encoders to obtain pan-knowledge representations while generating knowledge-specific representations based on injected adapters that are trained on an external knowledge source. With the specific knowledge, ConKI is able to produce more accurate sentiment predictions than solely using the pan-knowledge representations.
To further improve the learning of these representations, we specifically design a hierarchical contrastive learning procedure taking into account the contrasts between knowledge types within each modality, across modalities within one sample, and across samples. 
Experimental results on three benchmark datasets show that ConKI outperforms all state-of-the-art methods on a range of performance metrics.

\section*{Limitations}
Our research presents an initial step toward a knowledge injection framework for MSA and still has some limitations to be tackled in the future. 
Firstly, we can learn more disentangled representations by carefully selecting contrastive pairs for further improvement. 
Secondly, it will be interesting if we extend our method with multiple external sources that come from different knowledge domains.

\bibliography{anthology,custom}
\bibliographystyle{acl_natbib}

\appendix

\label{sec:appendix}

\section{Hierarchical Contrastive Learning}
\subsection{Pairing Policy} \label{pair}
\begin{figure}[ht]
\centering
  \includegraphics[width=\columnwidth]{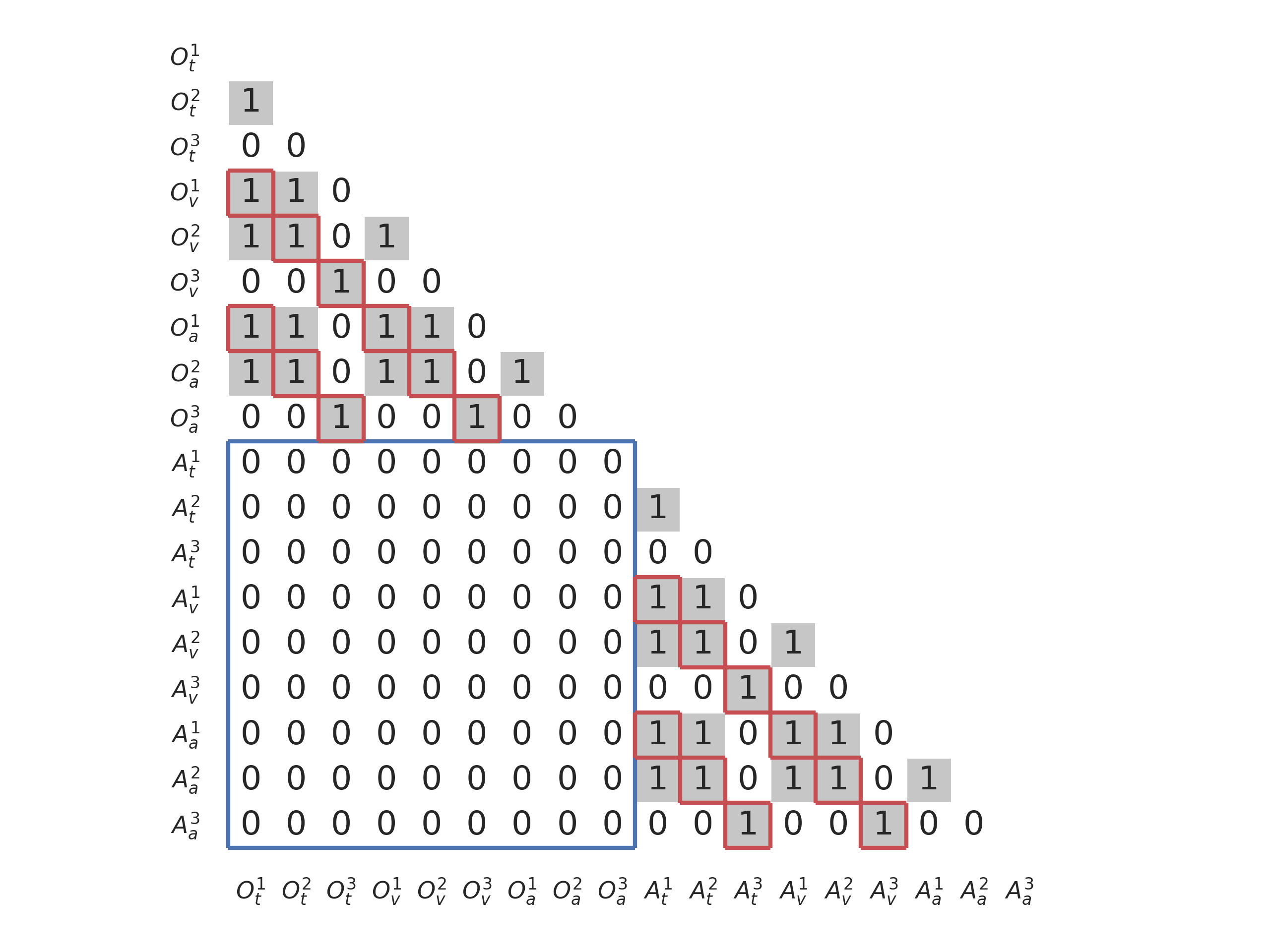}
  \caption{Pairing example of three samples where sample $1$ and sample $2$ are in the same sentiment interval while sample $3$ is in a different sentiment interval. Grey cells with ``1'' stand for the positive pairs, and white cells with ``0'' represent the negative pairs.} 
  \label{fig:cl_exam}
\end{figure}

To further elaborate on the pairing policy of our hierarchical contrastive learning, we show an example batch that consists of three samples ($\{O_t^i, O_v^i, O_a^i, A_t^i, A_v^i, A_a^i\}_{i=1, 2, 3} \in B$) where sample $1$ and sample $2$ belong to the same sentiment interval while sample $3$ falls in a different sentiment interval in Figure \ref{fig:cl_exam}. In this figure, the ``1''s in the heatmap represent the positive pairs of the row vectors and column vectors. The ``0''s represent the negative pairs of each two vectors. 
\begin{itemize}
    \item From this figure, we can get the \textit{intra-sample} positive pairs as the ``1''s with red borders:
            \begin{equation*}
            \begin{split}
                \mathcal{P}_1^B = \{(O_t^1, O_v^1), (O_t^1, O_a^1), (O_a^1, O_v^1), 
                \\(A_t^1, A_v^1), (A_t^1, A_a^1), (A_a^1, A_v^1),\\
                (O_t^2, O_v^2), (O_t^2, O_a^2), (O_a^2, O_v^2), 
                \\(A_t^2, A_v^2), (A_t^2, A_a^2), (A_a^2, A_v^2),\\
                (O_t^3, O_v^3), (O_t^3, O_a^3), (O_a^3, O_v^3), 
                \\(A_t^3, A_v^3), (A_t^3, A_a^3), (A_a^3, A_v^3)\};
            \end{split}
            \end{equation*}
    \item We represent the \textit{inter-knowledge} negative pairs $\mathcal{N}_1^B$ as the ``0''s in the blue zone; 
    \item Since sample $1$ and sample $2$ have close sentiment scores, we form the \textit{inter-sample} positive pairs as the ``1''s without red borders:
        \begin{equation*}
            \begin{split}
                \mathcal{P}_2^B = \{(O_t^1, O_t^2), (O_t^1, O_v^2), (O_t^1, O_a^2), 
                \\(O_v^1, O_t^2), (O_v^1, O_v^2), (O_v^1, O_a^2),\\ (O_a^1, O_t^2), (O_a^1, O_v^2), (O_a^1, O_a^2), \\(A_t^1, A_t^2), (A_t^1, A_v^2), (A_t^1, A_a^2),\\ (A_v^1, A_t^2), (A_v^1, A_v^2), (A_v^1, A_a^2),\\ (A_a^1, A_t^2), (A_a^1, A_v^2), (A_a^1, A_a^2)\};
            \end{split}
        \end{equation*}
        \item The remaining white cells with ``0'' show the negative pairs in $\mathcal{N}_2^B$ which aim to push sample $3$ away from sample $1$ and sample $2$ because they have different sentiments.
\end{itemize}

\subsection{Visualization of Modality Representations} \label{visual}
\begin{figure*}[t]
    \centering
  \includegraphics[width=\linewidth]{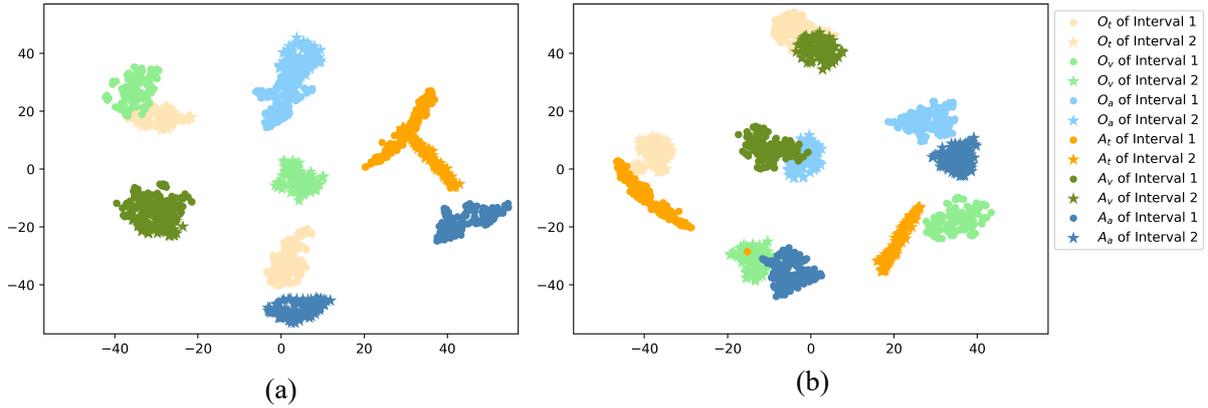}
  \caption{The visualization of the six decomposed representations of samples in the same sentiment interval and different intervals in (a) w/o h-CL; (b) ConKI. In each subfigure, light yellow, light blue, and light green represent pan-knowledge representations of text, audio, and video modalities, respectively, while dark yellow, dark blue, and dark green represent knowledge-specific representations accordingly. Each point or star stands for a sample in Interval 1 or Interval 2.} 
  \label{fig:tsne}
\end{figure*}
The motivation for us to propose hierarchical contrastive learning into ConKI is that we think modalities will be closer to each other within one sample and across samples in the same sentiment interval and will be far away across samples in different intervals while two knowledge contained in one modality will also be different. We use t-SNE \cite{van2008visualizing} to visualize the distributions of the six representations learned by ConKI with and without hierarchical contrastive learning, as shown in Figure \ref{fig:tsne}. 

Though we divide all samples into seven intervals to perform contrastive learning, we take samples of two intervals from the testing set to show the learned representations before and after contrastive learning due to the simplicity of the visualization.
From Figure \ref{fig:tsne}~(a), we can easily observe that some of the representations such as the pan-knowledge representations in light blue and the knowledge-specific representations in dark green of samples in two different intervals overlap extremely with each other. 

In contrast, these overlapping representations are pushed further in Figure \ref{fig:tsne}~(b) due to sample-level contrasts. It is also obvious that the three knowledge-specific representations of samples in the same interval, e.g., $A_t$, $A_v$, $A_a$ of Interval $2$ in dark colors and star shape become closer because of both modality-level and sample-level contrasts. Moreover, the distance between the knowledge-specific representations and the pan-knowledge representations, e.g., $A_v$ in dark green and $O_v$ in light green of Interval $2$, becomes larger in Figure \ref{fig:tsne}~(b) by knowledge-level contrasts. All of these indicate ConKI is able to perform desired contrastive learning for learning better representations that help improve the performance, even in the generalized scenario, i.e., in the testing set.  

\section{Implementation Details} \label{Imple_details}
We use unaligned raw data in all experiments as the previous works \cite{yu2021learning, han2021improving} for fair comparisons.
For audio and video modalities, two commonly-used toolkits (COVAREP \cite{6853739} and Facial Action Coding System (FACS) \cite{Ekman2005WhatTF}) act as the feature extractors, respectively. 
We use uncased 12-layers BERT pretrained model\footnote{\url{https://huggingface.co/bert-base-uncased}} as the text encoder, and two 2-layer transformers as the video and audio encoders, respectively. $\text{Adapter}_t$ has three modules inserted before the first, sixth, and eleventh layers of BERT sequentially. $\text{Adapter}_v$ and $\text{Adapter}_a$ have one module inserted before the second layer of the corresponding encoder.
We use CMU-MOSEI as the external dataset for CMU-MOSI and SIMS while using CMU-MOSI for CMU-MOSEI.
During the pretraining, the learning rate is set to $5e-5$ and we train for 10 epochs with one epoch for a linear warm-up scheduler. 
During the fine-tuning of CMU-MOSI and SIMS, the learning rates for encoders and other components are set to $5e-6$ and $1e-6$ respectively with a weight decay $0.001$. The temperature parameter $\tau$ is set to $0.07$ and $\lambda$ is set to $0.01$ after grid-search. We fine tune for 200 epochs with batch size 32. 
For the fine-tuning of CMU-MOSEI, the learning rates for the text encoder and others are $5e-6$ and $5e-5$, respectively. $\lambda$ for CMU-MOSEI is set to $0.001$.   
The best performance on the validation dataset is used for testing. 
We implement our experiments using PyTorch \cite{paszke2019pytorch} on an Nvidia RTX 2080Ti GPU.

\section{Datasets} \label{datasets}
CMU-MOSI is a popular benchmark dataset collected from YouTube. 
It contains 2,199 video clips sliced from 93 videos where a speaker shares opinions on topics such as movies. Each video is annotated with sentiment scores ranging from $-3$ (strongly negative) to $+3$ (strongly positive).
CMU-MOSEI is the largest MSA dataset that has greater diversity in speakers, topics, and annotations. It contains 22,856 annotated video segments from 1,000 distinct speakers and 250 topics. Each clip also has sentiment scores between $[-3, +3]$. 
SIMS is a Chinese MSA dataset that contains 2,281 refined video segments. Each sample has one multimodal label and three unimodal labels, with sentiment scores from $-1$ to $+1$. We translate the Chinese text into English\footnote{\url{https://pypi.org/project/googletrans/}} so that we can inject knowledge from English MSA datasets into SIMS. For fair comparisons, all baseline models use the English version to evaluate the performance. 

\section{Baseline Models} \label{baselines}
\textbf{TFN.} The Tensor Fusion Network (TFN) \cite{zadeh2017tensor} encodes three modalities with corresponding embedding subnetworks and uses outer-product to model the unimodal, bimodal, and trimodal interactions as the fusion results. 

\textbf{LMF.} The Low-rank Multimodal Fusion (LMF) \cite{liu2018efficient} utilizes low-rank tensors to improve efficiency of multimodal fusion.

\textbf{MulT.} The Multimodal Transformer (MulT) \cite{tsai2019multimodal} proposes directional pairwise cross-modal attention that adapts one modality into another for multimodal fusion.

\textbf{ICCN.} The Interaction Canonical Correlation Network (ICCN) \cite{sun2020learning} learns text-based audio and text-based video features by optimizing canonical loss. These features are concatenated with the text features for downstream classifiers such as logistic regression.

\textbf{MISA.} The Modality-Invariant and -Specific Representations (MISA) \cite{hazarika2020misa} designs a multitask loss including task prediction loss, reconstruction loss, similarity loss, and difference loss to learn modality-invariant and modality-specific representations. 

\textbf{MAG-BERT.} The Multimodal Adaptation Gate for Bert (MAG-BERT) \cite{rahman2020integrating} builds an alignment gate that allows audio and video information to leak into the BERT model for multimodal fusion.

\textbf{Self--MM.} The Self-Supervised Multitask Learning (Self--MM) \cite{yu2021learning} proposes a label generation module based on self-supervised learning to obtain unimodal supervision. Then they joint train the multimodal and unimodal tasks for better fusion results.

\textbf{HyCon.} The Hybrid Contrastive Learning (HyCon) \cite{mai2021hybrid} performs intra- and inter-modal contrastive learning as well as semi-contrastive learning within a modality to explore cross-modal interactions.

\textbf{MMIM.} MultiModal InfoMax (MMIM) \cite{han2021improving} maximizes the mutual information in unimodal input pairs as well as between multimodal fusion result and unimodal input to aid the main MSA task.

\end{document}